\documentclass[letterpaper]{article} 
\usepackage{aaai24}  
\usepackage{times}  
\usepackage{helvet}  
\usepackage{courier}  
\usepackage[hyphens]{url}  
\usepackage{graphicx} 
\usepackage{natbib}  
\usepackage{caption} 
\usepackage{algorithm}
\usepackage{algorithmic}

\usepackage{amsmath,amssymb,amsfonts}
\usepackage{amsthm}
\usepackage{float}
\usepackage{enumitem}
\usepackage{booktabs}
\usepackage{amssymb}
\usepackage{colortbl}
\usepackage{graphicx}
\usepackage{multirow}
\usepackage[table]{xcolor}
\usepackage[switch]{lineno}
\usepackage{bbding}
\usepackage{booktabs}
\usepackage{soul}
\usepackage{subfig}

\usepackage{newfloat}
\usepackage{listings}
\DeclareCaptionStyle{ruled}{labelfont=normalfont,labelsep=colon,strut=off} 
\floatstyle{ruled}
\newfloat{listing}{tb}{lst}{}
\floatname{listing}{Listing}
\title{Efficient Prompt Tuning by
Multi-Space Projection and Prompt Fusion}
\author{
    Pengxiang Lan, Enneng Yang, Yuting Liu, Guibing Guo\thanks{Corresponding Author.}, Jianzhe Zhao, Xingwei Wang\footnotemark[1]
}
\affiliations{
    Northeastern University, China\\

    \{pengxianglan, ennengyang, yutingliu\}@stumail.neu.edu.cn, \{guogb, zhaojz\}@swc.neu.edu.cn, wangxw@mail.neu.edu.cn
}

\usepackage{bibentry}

\begin{document}

\maketitle

\begin{abstract}
Prompt tuning is a promising method to fine-tune a pre-trained language model without retraining its large-scale parameters. Instead, it attaches a soft prompt to the input text, whereby downstream tasks can be well adapted by merely learning the embeddings of prompt tokens. Nevertheless, existing methods still suffer from two challenges: \textbf{(i)} they are hard to balance accuracy and efficiency. A longer (shorter) soft prompt generally leads to a better (worse) accuracy but at the cost of more (less) training time.  \textbf{(ii)} The performance may not be consistent when adapting to different downstream tasks. We attribute it to the same embedding space but responsible for different requirements of downstream tasks. To address these issues, we propose an \textbf{E}fficient \textbf{P}rompt \textbf{T}uning method (\textbf{EPT}) by multi-space projection and prompt fusion. Specifically, it decomposes a given soft prompt into a shorter prompt and two low-rank matrices, significantly reducing the training time. Accuracy is also enhanced by leveraging low-rank matrices and the short prompt as additional knowledge sources to enrich the semantics of the original short prompt. In addition, we project the soft prompt into multiple subspaces to improve the performance consistency, and then adaptively learn the combination weights of different spaces through a gating network. Experiments on 13 natural language processing downstream tasks show that our method significantly and consistently outperforms 11 comparison methods with the relative percentage of improvements up to 12.9\%, and training time decreased by 14\%.
\end{abstract}
\section{Introduction}
Fine-tuning methods have become a growing focus to adapt a pre-trained language model (PLM) to a variety of downstream tasks \citep{devlin2019bert,radford2019language}. However, the continuous expansion of the PLMs scale has led to a significant increase in the number of parameters \citep{zhang2022opt}, such as the T5 model \citep{raffel2020exploring} containing hundreds of millions of parameters. Therefore, full fine-tuning PLMs on all parameters is unrealistic in practical applications. The discrete phrase-based tuning provides task descriptions in the form of input text \citep{brown2020language}, guiding PLMs to perform corresponding downstream tasks effectively, avoiding full-parameter fine-tuning. Unfortunately, manually designing an effective set of task prompt phrases heavily relies on experts' domain knowledge, which is still challenging in the face of a wide variety of tasks. 
\begin{figure}[t]
\centering
  \includegraphics[scale=0.25]{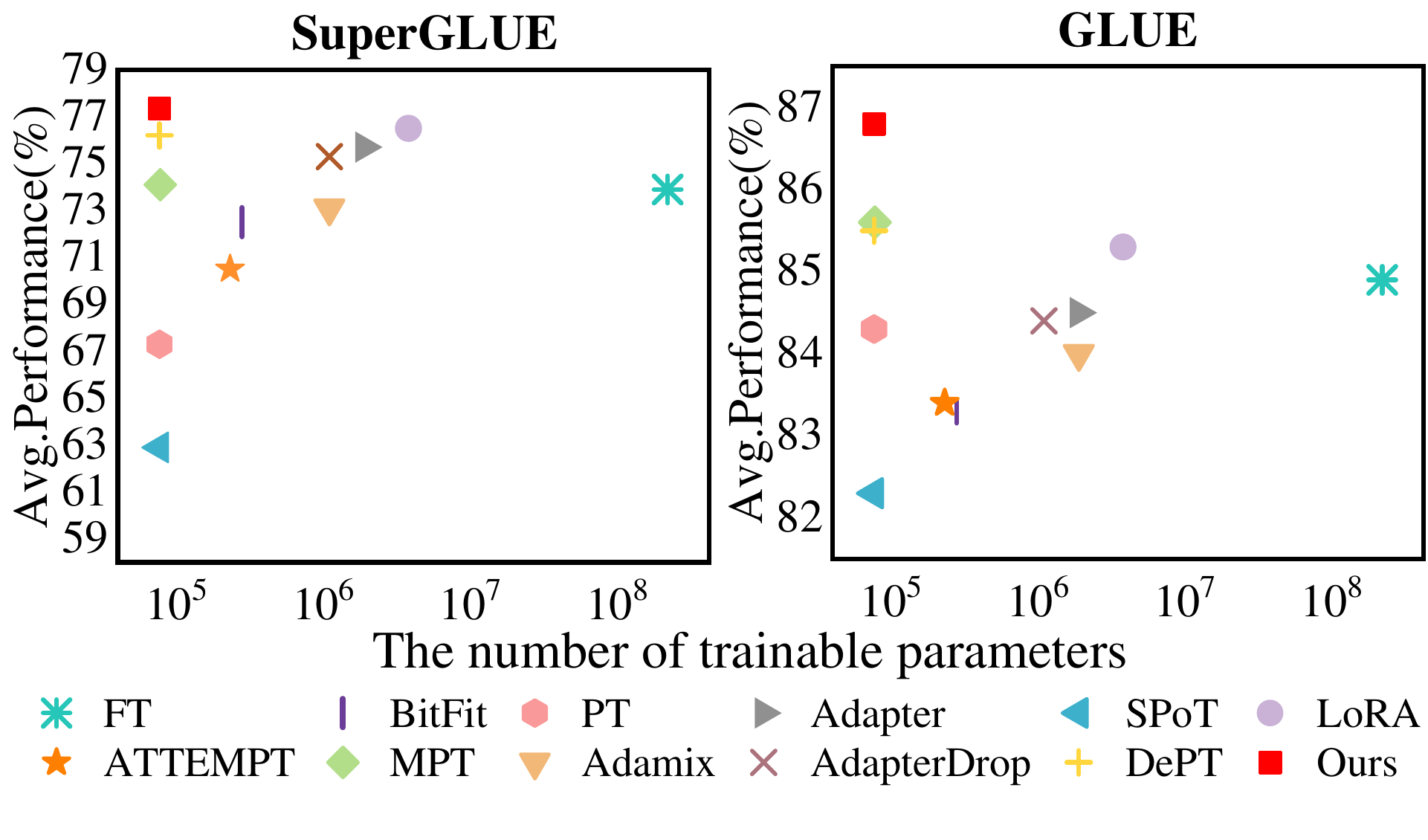}
    \caption{Average performance ($y$-axis) against the number of trainable parameters ($x$-axis) on the GLUE and SuperGLUE benchmarks. We utilize the T5-Base for all models. 
    }
    \label{Figure.1}
\end{figure}

Recently, prompt tuning (PT) \citep{lester2021power} based method has become an effective alternative to convert discrete phrases into a set of learnable parameters. PT freezes the parameters of PLMs and only trains the attached soft (continuous) prompt vectors to the input text. Therefore, its parameters do not dramatically scale up with the expansion of the model size, making PT stand out in the parameter-efficient fine-tuning (PEFT) approaches \citep{shi2024dept}. Recent studies have leveraged some successful approaches to reduce training parameters in PT, such as parameter-efficient transfer learning (PETL) \citep{vu2022spot,asai2022attempt}, multi-task learning \citep{wang2022multitask}, and decomposing soft prompts \citep{shi2024dept,xiao2023decomposed}. 
Despite these PT variants effectively improving soft prompt performance in downstream tasks, PT still faces several limitations that cannot be ignored. $\textbf{First}$, existing PT-based methods encounter the challenge of balancing accuracy and efficiency \citep{xiao2023decomposed,lester2021power, shi2024dept}. Attaching the soft prompt to the input extends the overall length of the input sequence. Due to the quadratic complexity of the Transformer \citep{vaswani2017attention}, lengthening the soft prompt introduces additional training time. PT requires training a substantial number of prompt tokens to achieve competitive performance \citep{lester2021power}; directly shortening the soft prompt to reduce training time may result in sub-optimal performance for PT. $\textbf{Second}$, existing PT-based variants are not well adapted to various downstream tasks and are causing inconsistent performance. This is because they attempt to handle the different needs of various downstream tasks with the same embedding space \citep{shi2024dept,wang2022multitask,asai2022attempt}. However, text information in natural language processing tasks involves different types \citep{wang2019superglue} and degrees of difficulty, and models pay limited attention to semantics in the short prompt. For example, on the SuperGLUE \citep{wang2019superglue} benchmark, which is more complex than the GLUE \citep{wang2018glue} benchmark, the performance of PT's variants is not very satisfactory.

To tackle the aforementioned knotty issues, we propose a novel efficient prompt tuning (EPT) that consists of two core modules: prompt fusion and multi-space projection. EPT initially decomposes a whole soft prompt into two independent parts: a short prompt and two low-rank matrices. Only the short prompt is attached to the front of the input, to reduce the training time. Low-rank matrices are utilized to update the frozen input text embedding. Next, to offset the semantic loss of the short prompt compared with long ones, we design a prompt fusion module. This module utilizes the attention network by Einstein Summation to capture the knowledge difference between low-rank matrices and the short prompt, and instills this difference into the short prompt to improve the semantic richness of the short prompt. Then, to adapt PT to different downstream tasks more consistently, we leverage a multi-space projection module to project a single soft prompt into multiple subspaces and reweight the soft prompt in these subspaces according to the task through the gating network. Finally, a joint representation of the prompt (obtained from the prompt fusion and multi-space modules) replaces the vanilla prompt.

\ \\ \noindent \textbf{Contributions.} In summary, the main contributions of this paper are as follows:
\begin{itemize}
\item We point out that PT-based methods suffer from the trade-off dilemma of ``accuracy and efficiency'' as well as performance inconsistency. To address these issues, we propose a novel efficient prompt tuning (EPT) method.

\item We design two effective modules in EPT, prompt fusion and prompt projection. The former helps to maintain the efficiency of the short prompt and compensate for the semantic missing of the short prompt to enhance performance, and the latter reweights prompts in multiple subspaces to adapt to downstream tasks.

\item We comprehensively evaluate EPT on the GLUE and SuperGLUE benchmarks, where EPT outperformed other PEFT methods, including LoRA and multi-task transfer learning-based PT variants (see Figure. \ref{Figure.1}). In particular, EPT achieves a 14\% reduction in training time compared to vanilla PT on the GLUE benchmark.
\end{itemize}

\begin{figure}[t]
\centering
  \includegraphics[scale=0.81, trim={0mm 0mm 0mm 0mm}]{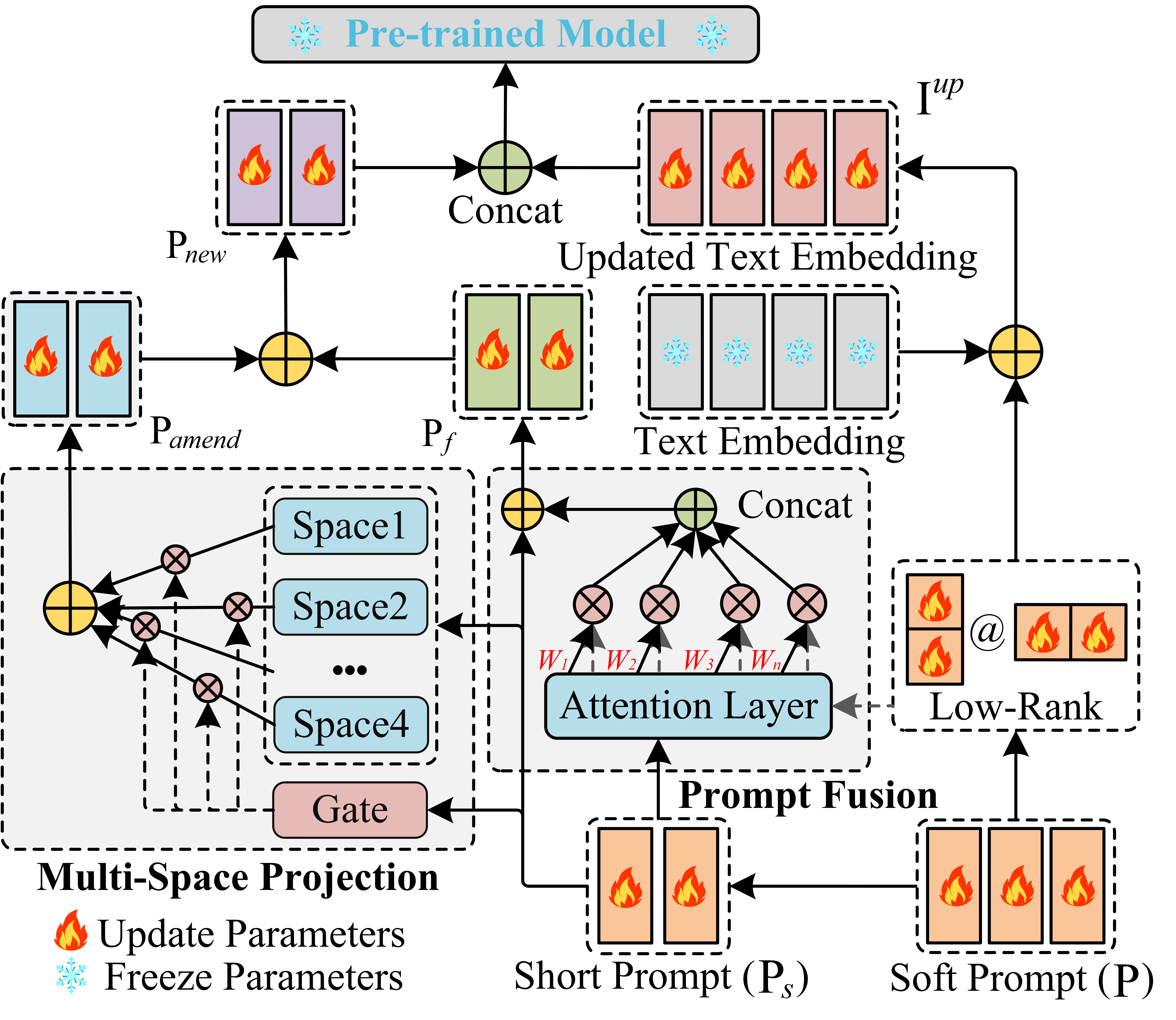}
    \caption{
    The overview of the EPT model. The whole soft prompt is decomposed into a short prompt and two low-rank matrices. Low-rank matrices are multiplied and added element-wise to the frozen input text embedding. The Multi-Space Projection Module maps the short prompt to multiple subspaces, addressing diverse downstream task requirements, while the Prompt Fusion module enhances its semantic knowledge. Finally, EPT generates a joint prompt representation to supersede the original prompt. The new prompt and the updated input text embedding are concatenated to input into the PLM. 
    }
    \label{Figure.2}
\end{figure}

\section{The Proposed Method}
In this section, we first introduce the background of the prompt tuning and then elaborate our proposed EPT method as shown in Figure.~\ref{Figure.2}. It consists of four main modules: (1) Prompt Decomposition, (2) Prompt Fusion, (3) Multi-Space Projection,  and (4) Reconstructed Prompt. 

\subsection{Background: Prompt Tuning}
We first introduce the training method of PT. PT has gained widespread adoption in downstream tasks due to its advantage of the parameters not increasing sharply with the expansion of the model \cite{shi2024dept}. Let labeled training data $(\boldsymbol{X}, \boldsymbol{Y})=\left\{\boldsymbol{x}_i, \boldsymbol{y}_i\right\}_{i=1}^N$ for one target task $\mathcal{T}$, where $N$ is the number of training data. Given a PLM with parameters $\Theta$ and each input text $x_{i}$. The embedding of $x_{i} \in \boldsymbol{X}$ is represented as $\mathbf{E}_{i}\in \mathbb{R}^{{m \times d}}$, where $m$ is maximum sequence length and $d$ is the hidden dimension of input text embedding. $\mathbf{P} \in \mathbb{R}^{{l \times d}}$ is initialized to form a target prompt, $l$ is a hyper-parameter for the length of the soft prompt. It is concatenated with $\mathbf{E}_{i}\in \mathbb{R}^{{m \times d}}$, which does not involve gradient updates during training, to form a new input embedding $\left [ \mathbf{P};\mathbf{E}_{i} \right]\in \mathbb{R}^{(l+m)\times d}$. The target task is formulated as follows:
\begin{equation}
    \mathcal{L}_{PT} = -\sum_{i}\log P \left(\mathbf{y}_{i}|\left[\mathbf{P};\mathbf{E}_{i} \right];\Theta \right)
\end{equation}
where $\mathcal{L}_{PT}$ is a loss function only optimized with the prompt $\mathbf{P}$. However, the vanilla PT requires training a large number of prompt tokens (i.e., a larger value of $l$ in $\mathbf{P}$) to achieve the expected performance \cite{lester2021power}.

\subsection{Prompt Decomposition}
Most studies have shown that the performance of PT is comparable to full fine-tuning \citep{razdaibiedina-etal-2023-residual,wang2022multitask}. 
However, a challenging issue persists: PT requires training a substantial number of prompt tokens to achieve competitive performance, resulting in an increased length of the entire input sequence \citep{lester2021power}. It causes greater resource consumption in the training/inference phase. We begin by initializing our source prompt $\mathbf{P}\in \mathbb{R}^{l\times d}$ from sampled vocabulary (e.g., the 5000 most common tokens) to ensure that $\mathbf{P}$ is informative content. Inspired by DEPT \citep{shi2024dept}, we truncate a trainable short prompt $\mathbf{P}_{s}\in \mathbb{R}^{s\times d}$ with a length of $s$ ($s < l$) from $\mathbf{P}$. Subsequently, we align the dimensions of $\mathbf{P}\in \mathbb{R}^{l\times d}$ with $\mathbf{E}\in \mathbb{R}^{{m \times d}}$ and then perform Singular Value Decomposition (SVD), retaining the top $r$ two trainable low-rank singular vector matrices ( $\mathbf{A}\in \mathbb{R}^{m\times r}$ and $\mathbf{B}\in \mathbb{R}^{r\times d}$). Among them, $r$ is the rank in low-rank matrices and $r \ll \min(m,d)$, 
$d$ is the dimension of input text embedding, $m$ is the maximum sequence length. Due to the transformer's quadratic complexity, the training duration is proportional to the length of the prompt. Therefore, a shorter prompt $\mathbf{P}_{s}$ can effectively reduce the training time. Notably, unlike the DEPT, its $\mathbf{A}$ and $\mathbf{B}$ are random Gaussian initialization and zero initialization respectively (follow LoRA \citep{hu2021lora}). This operation of randomly initializing results in a complete loss of information about the original longer prompt, $\mathbf{P}$, since it is semantically rich. Therefore, in our approach, $\mathbf{A}$ and $\mathbf{B}$ are obtained by decomposing of $\mathbf{P}$ to preserve the semantic knowledge of original prompt $\mathbf{P}$ as much as possible.

To keep the same amount of trainable parameters, the selection of $s$ and $r$ satisfies the equation $l\times d=s\times d+(m+d)\times r$, where $s$ and $r$ are hyper-parameters and $s<l$ when $r>0$. For the decomposition of the vanilla PT, the specific values of $s$ and $r$ affect each other. For example, in the T5-base, $d$ (dimension) is 768. If $l$ is 100 and $m$ is 256, when the length of $\textbf{P}_{s}$ is 60, $r$ is 30 ($60 \times 768+(256 +768) \times 30$). When the length of $\textbf{P}_{s}$ is 40, $r$ is 45 ($40 \times 768+(256 +768) \times 45$). When $r=0$, $s=l$, the decomposed PT proposed in this paper degenerates to vanilla PT. The purpose of the low-rank matrices is to update the frozen input word embedding. When $s=0$, only low-rank matrices are used to update the frozen input word embedding:
\begin{eqnarray}
    \mathbf{I}^{up}_{i}=\mathbf{E}_{i}+\mathbf{A}\otimes\mathbf{B}
\end{eqnarray}
where $\mathbf{A}\otimes\mathbf{B}$ represents the multiplication operation of $\mathbf{A}$ and $\mathbf{B}$, $\mathbf{I}^{up}_{i}\in \mathbb{R}^{m\times d}$ represents the result of adding $\mathbf{A}\otimes\mathbf{B}$ to the frozen input text embedding $\mathbf{E}_{i}$.

\subsection{Prompt Fusion}
\label{subsec:promptfusion}
In this section, we design a novel prompt fusion module to keep the short prompt efficiency and further compensate for the semantic loss of the decomposition of the long prompt into a short prompt and two low-rank matrices in the previous section. Specifically, supposing the short prompt $\mathbf{P}_{s}$ is directly injected into PLMs (the vanilla prompt has the same operation). In that case, although shortening the length of the soft prompt reduces the training time, this still will lead to poor performance of PT because of the lack of knowledge of the original prompt $\mathbf{P}$. PT requires a substantial number of prompt tokens (exceeding 100) to achieve optimal performance \citep{lester2021power}. Therefore, enriching the knowledge of $\mathbf{P}_{s}$ becomes exceptionally crucial while reducing the training time. 

Building upon this foundation, we first leverage an attention network by Einstein Summation to consider the difference in knowledge richness between low-rank matrices and the short prompt. Then, we add the short prompt with the output of the attention network to enhance the knowledge of the original short prompt:
\begin{eqnarray}
   \textbf{W}_{attn} = \mbox{softmax}(\frac{1}{\sqrt{d}} \mathbf{P}_{s} \cdot (\mathbf{A}\otimes\mathbf{B})^{\top})
\end{eqnarray}
\begin{eqnarray}
    \mathbf{P}_{f} = \mathbf{P}_{s}+ Ein(\textbf{W}_{attn} \cdot \mathbf{P}_{s}) 
\end{eqnarray}
where $(\mathbf{A}\otimes\mathbf{B})^{\top}$ is the transpose of $\mathbf{A}\otimes\mathbf{B}$, $\textbf{W}_{attn}$ is the weighted vector representation, and $Ein(\cdot)$ is the Einstein Summation (the way the dimensions change is $'bpl,bpd \rightarrow bpd'$). The attention mechanism $\textbf{W}_{attn}\cdot \mathbf{P}_{s}$ considers the knowledge association between low-rank matrices and $\mathbf{P}_{s}$. $\mathbf{P}_{f} \in \mathbb{R}^{m\times d}$ enhances the knowledge within the original short prompt based on reducing the consumption of computing resources. 

\subsection{Multi-Space Projection}
\label{subsec:multispace}
In this section, we propose the multi-space projection module to project a single prompt into multiple subspaces to solve the performance inconsistency problem of the original PT only fine-tuning in a single space, which reweights the prompt representations in different spaces through a gating network at each downstream task. Text information in text classification tasks usually involves different types and degrees of difficulty (such as Natural Language Inference, Question Answering, etc.). However, PT is inputted into PLMs in the same embedding space to adapt to downstream tasks, and a single space does not consider the different requirements in downstream tasks. This results in potentially inconsistent performance of PT - as it performs well on some tasks and poorly on others. The Mixture-of-Experts ~\citep{jacobs1991adaptive} provides an excellent idea to solve the aforementioned problem. Motivated by this, we map $\mathbf{P}_{s}$ to distinct spaces and utilize a gating network to control each space's weight distribution. Prompt tokens are assigned different degree weights by achieving the parameter selection:
\begin{eqnarray}
    E_{i}(\mathbf{P}_{s}) = \mbox{linear}_{1} \left(\sigma \left(\mbox{linear}_{2}\left(\mathbf{P}_{s} \right)\right)\right), i \in [1,...,N_e]
\end{eqnarray}
where $E_{i}(\mathbf{P}_{s}) \in \mathbb{R}^{s\times d}$ is the $i$-th space, $\mbox{linear}_{1} \in \mathbb{R}^{m\times d}$, $\mbox{linear}_{2} \in \mathbb{R}^{d\times m}$, $N_e$ is the maximum number of spaces, the activation function $\sigma(\cdot)$ is a ReLU~\citep{krizhevsky2012imagenet} function. The gate network is formulated as follows:
\begin{eqnarray}
    \begin{split}
    &f_{i}(\mathbf{P}_{s})=\mbox{linear}(\mathbf{P}_{s}), i \in [1,...,N_e] \\
    &G_{i}(\mathbf{P}_{s}) = \frac{\mbox{exp}^{f_{i}(\mathbf{P}_{s})}}{\sum^{N_e}_{i=1} \; \mbox{exp}^{f_{i}(\mathbf{P}_{s})}}
    \end{split}
\end{eqnarray}
where $G_{i}(\mathbf{P}_{s}) \in \mathbb{R}^{s \times 1}$ is used to control the importance of each space, $\mbox{linear} \in \mathbb{R}^{d\times 1}$. Reweighting each space by leveraging a gating mechanism:
\begin{eqnarray}
    \mathbf{P}_{amend}=\sum_{i=1}^{N_e} \; G_{i}(\mathbf{P}_{s})\cdot E_{i}(\mathbf{P}_{s})
\end{eqnarray}
where $\mathbf{P}_{amend} \in \mathbb{R}^{s \times d}$ is the result of reweighting $\mathbf{P}_{s}$. $G_{i}(\mathbf{P}_{s})$ makes one or more spaces in an active state better for different parameter selections.

\subsection{Reconstructed Prompt}
In this section, our EPT method integrates prompt representations of the fusion module and the multi-space module to obtain a joint representation to have both advantages.
To be specific, we learn the joint representation $\mathbf{P}_{new}$ of $\mathbf{P}_{amend}$ and $\mathbf{P}_{f}$. Weights of $\mathbf{P}_{amend}$ are allocated in different spaces, and the soft prompt $\mathbf{P}_{f}$ in the prompt fusion module. The purpose of learning a joint representation of soft prompts is to replace the original prompt $\mathbf{P}$ with $\mathbf{P}_{new}$:
\begin{eqnarray}
    \mathbf{P}_{new} = \mathbf{P}_{amend}+\mathbf{P}_{f}
\end{eqnarray}
when the initialized $\mathbf{P}_{s}$ performs poorly on specific tasks, $\mathbf{P}_{amend}$ and $\mathbf{P}_{f}$ redistribute the importance of $\mathbf{P}_{s}$. After learning $\mathbf{P}_{new}$, the constructed network is discarded, and $\mathbf{P}_{new}$ is utilized for training in the PLM. Therefore, the trainable parameters input into the PLMs will remain consistent with the original PT. By $\mathbf{P}_{new}$ and $\mathbf{I}^{up}_{i}$, Eq.(1) is displaced by:
\begin{eqnarray}
\mathcal{L}_{PT} = -\sum_{i}\; \log P(\mathbf{y}_{i}|[\mathbf{P}_{new};\mathbf{I}^{up}_{i}];\mathbf{P}_{new})
\end{eqnarray}
where $[\mathbf{P}_{new};\mathbf{I}^{up}_{i}]$ is a input embedding of PLMs through the connection of $\mathbf{P}_{new}$ and $\mathbf{I}^{up}_{i}$.

\subsection{Quantization}
To reduce GPU memory usage, we employed quantization techniques \citep{dettmers8, dettmers2023qlora} for models with a size of 3B or larger. This process involves rescaling the input tensors by loading the model in 4-bit precision and back-quantizing the values to bf16 during training. We minimize storage consumption by implementing the double quantization method proposed in QLoRA \citep{dettmers2023qlora}, which approach significantly reduces memory usage while maintaining performance comparable to standard parameter-efficient fine-tuning. Notably, weight gradients are still calculated exclusively on the soft prompt parameters.

\section{Experiments}
We conduct extensive experiments to answer these key research questions: 
\textbf{RQ1:} How does EPT compare with state-of-the-art baselines across different datasets? 
\textbf{RQ2:} How do we understand the impact of the critical components of EPT and model scaling on the performance of EPT? 
\textbf{RQ3:} How do the few-shot adaptability and hyper-parameter tuning affect the performance of EPT?
\subsection{Evaluation Datasets and Source Tasks}
We conducted multi-angle experiments on the EPT method to demonstrate its outstanding applicability to 13 publicly available NLP tasks (8 from the GLUE benchmark \footnote{\footnotesize \url{https://huggingface.co/datasets/glue}} and 5 from the SuperGLUE benchmark \footnote{\footnotesize \url{https://huggingface.co/datasets/super_glue}}). Specifically, \textbf{(1) GLUE} ~\citep{wang2018glue} is a benchmark for evaluating natural language understanding performance. It consists of diverse tasks that test the model's ability to understand language in different contexts. To fully prove the performance effect of EPT, we maintain consistency with previous work, and the NLP datasets are MNLI ~\citep{williams2018broad}, QQP ~\citep{wang2018glue}, QNLI ~\citep{rajpurkar2016squad}, SST-2 ~\citep{socher2013recursive}, STS-B ~\citep{cer2017semeval}, MRPC ~\citep{dolan2005automatically}, RTE ~\citep{giampiccolo2007third} and CoLA ~\citep{warstadt2019neural} from GLUE. 
\textbf{(2) SuperGLUE} ~\citep{wang2019superglue} is an extension of GLUE, that includes more complex and challenging tasks.  
This paper uses five tasks from SuperGLUE: MultiRC ~\citep{khashabi2018looking}, BoolQ ~\citep{clark2019boolq}, WiC ~\citep{pilehvar2019wic}, WSC ~\citep{levesque2012winograd} and CB ~\citep{de2019commitmentbank}. We follow the previous working setup ~\citep{su2022transferability,asai2022attempt,shi2024dept}, which only utilizes ReCoRD \citep{zhang2018record} and SQuAD \citep{rajpurkar2016squad} in the few-shot experiment. 
Appendix A shows the complete statistics of all experimental datasets.

%
\begin{table*}[t]
\centering
\renewcommand{\arraystretch}{0.9}
\setlength{\tabcolsep}{0.038cm}
\centering
\begin{tabular}{l|c|cccccccc|>{\columncolor{gray!25}}c|ccccc|>{\columncolor{gray!25}}c}
\hline
& & \multicolumn{9}{c}{\textbf{GLUE}} & \multicolumn{6}{c}{\textbf{SuperGLUE}}  \\ 
\cline{3-17}
  & & MNLI & QQP & QNLI & MRPC & STS-B & SST-2 & CoLA & RTE & \multirow{-1}{*}{Mean} & Multi & WiC & WSC & BoolQ & CB & \multirow{-1}{*}{Mean}  \\ 
\multirow{-3}{*}{\textbf{Model}} &\multirow{-3}{*}{Param} & (393K) &(364K) &(105K) & (3.7K) &(7K) &(67K) &(8.5K) &(2.5K) & (\%) & (5.1K) & (6K) & (554) & (9.4K) & (250) & (\%) \\
\hline
Fine-tuning$^1$ & 220M & \textbf{86.8} & \textbf{91.6} & 93.0 & \underline{90.2} & 89.7 & \textbf{94.6} & 61.8 & 71.9 & 84.9  & 72.8 & \textbf{70.2} & 59.6 & 81.1 & 85.7 & 73.9   \\ 
LoRA$^2$ & 3.8M & 86.3 & 89.0 & \underline{93.2} & 90.1 & 90.9 & 94.3 & 63.3 & 75.5 & 85.3  & 72.6 & 68.3 & \underline{67.3} & 81.3 & \textbf{92.9} & \underline{76.5} \\
Adapter$^1$ & 1.9M & \underline{86.5} & 90.2 & \underline{93.2} & 85.3 & 90.7 & 93.8 & 64.0 & 71.9 & 84.5  & \textbf{75.9} & 67.1 & \underline{67.3} & \textbf{82.5} & 85.7 & 75.7 \\
Adamix & 1.9M & 86.4 & 90.1 & 93.0 & 87.4 & 91.0 & 93.9 & 59.2 & 70.8 & 84.0  & 73.1 & 66.8 & 59.3 & 80.6 & 85.7 & 73.1  \\
AdapterDrop$^1$ & 1.1M & 86.3 & 90.2 & \underline{93.2} & 86.3 & \textbf{91.4} & 93.6 & 62.7 & 71.2 & 84.4  & 72.9 & 68.3 & \underline{67.3} & \underline{82.3} & 85.7 & 75.3   \\
BitFit$^1$ & 280K & 85.3 & 90.1 & 93.0 & 86.8 & 90.9 & 94.2 & 58.2 & 67.6 & 83.3 & 74.5 & \underline{70.0} & 59.6 & 79.6 & 78.6 & 72.5 \\
PT & 76.8K & 83.6 & 90.3 & 93.1 & 87.7 & 90.2 & 93.6 & 59.5 & 76.2 & 84.3 & 67.3 & 60.5 & 59.6 & 70.7 & 78.6 & 67.3\\
ATTEMPT$^{\bigstar1}$ & 232K & 84.3 & 90.3 & 93.0 & 85.7 & 89.7 & 93.2 & 57.4 & 73.4 & 83.4 & 74.4 & 66.8 & 53.8 & 78.8 & 78.6 & 70.5 \\
MPT$^{\bigstar3}$ & 77.6K & 85.9 & 90.3 & 93.1 & 89.1 & 90.4 & 93.8 & 62.4 & 79.4 & \underline{85.6} & \underline{74.8} & 69.0 & \underline{67.3} & 79.6 & 79.8 & 74.1 \\
SPoT$^{\bigstar1}$ & 76.8K & 85.4 & 90.1 & 93.0 & 79.7 & 90.0 & 93.4 & 57.1 & 69.8 & 82.3 & 74.0 & 67.0 & 50.0 & 77.2 & 46.4 & 62.9 \\
DEPT & 76.8K & 85.1 & \underline{90.4} & \textbf{93.3} & 89.2 & 91.0 & 94.2 & 62.7 & 78.4 & 85.5 & 74.4 & 67.1 & \underline{67.3} & 79.4 & \textbf{92.9} & 76.2 \\
DPT & 9.0K & 85.4 & 90.2 & 93.1 & \textbf{90.4} & 90.3 & 94.5 & 57.8 & 79.0 & 85.1 & 74.0 & 68.5 & \underline{67.3} & 79.4 & 78.6 & 73.6 \\
\hline
\textbf{EPT (ours)} & 76.8K & 85.8 & 90.3 & \underline{93.2} & \underline{90.2} & \underline{91.1} & \underline{94.5} & \textbf{67.0} & \underline{82.0} & \textbf{86.8} & \underline{74.8} & 69.0 & \textbf{69.2} & 81.5 & \textbf{92.9} & \textbf{77.5}\\
\hline
ATTEMPT$^{\lozenge\bigstar3}$  &96.0K & 83.7 & 90.1 & \underline{93.2} & 87.3 & 90.8 & 94.3 & \underline{64.3} & \textbf{82.7} & 85.8 & 74.4 & 66.5 & \textbf{69.2} & 78.5 & 82.1 & 74.1 \\
MPT$^{\lozenge\bigstar3}$ & 10.5K & 84.3 & 90.0 & 93.0 & 89.2 & 90.4 & 93.3 & 63.5 & \textbf{82.7} & 85.8 & \underline{74.8} & \textbf{70.2} & \underline{67.3} & 79.2 & 89.3 & 76.1\\
\hline
\end{tabular}
\caption{Performance comparison on GLUE and SuperGLUE benchmark, all experimental results are based on the T5-Base model. The evaluation metrics are Pearson correlation for STS-B,  F1 for MultiRC (Multi) and accuracy for other tasks. ``Param" represents the amount of trainable parameters for each task. Where $\bigstar$ indicates that some tasks utilize the PETL method, $\lozenge$ indicates that some tasks utilize multi-task learning (resulting in the reduction of trainable parameters). $^1$ sourced from \citep{asai2022attempt}. $^2$ sourced from \citep{sung2022lst}. $^3$ sourced from \citep{wang2022multitask}. The best result is marked in bold. The second-best result is marked with an underline. The numbers under datasets refer to training examples in each dataset.}
\label{tab_glue}
\end{table*}

\subsection{Baselines}
We focus on exploring a high-performance and less training parameter method of PEFT, so the number of training parameters is also an essential factor. Methods such as KronA \citep{edalati2022krona}, S4 \citep{chen2022parameter}, etc. have more training parameters, for example, the training parameter of PT is 0.1\% of full fine-tuning, while the training parameter of MAM adapter \citep{he2021towards} is 6.7\% of full fine-tuning. Therefore, we focus more on the latest methods of PT-type in the baseline selection.

The baselines for comparison with EPT are: \textbf{(1) Full Fine-tuning (FT)}, which updates all parameters of PLMs. \textbf{(2) PEFT approaches}, including Adapter ~\cite{houlsby2019parameter}, AdapterDrop ~\cite{ruckle2021adapterdrop}, AdaMix \citep{wang2022adamix}, BitFit ~\cite{zaken2022bitfit}, and LoRA \cite{hu2021lora}. \textbf{(3) PT-based method}, where the vanilla PT ~\cite{lester2021power} updates parameters with prompt prefix to accommodate various downstream tasks, and its variants include SPoT ~\cite{vu2022spot}, ATTEMPT ~\cite{asai2022attempt}, MPT ~\cite{wang2022multitask}, and their transfer and multi-task learning variants. SPoT and ATTEMPT find optimal prompt initializations by pre-training prompts on informative source tasks. \textbf{(4) Prompt decomposition}, DEPT~\citep{shi2024dept} and DPT ~\citep{xiao2023decomposed} are parameter-efficient method that decomposes the soft prompt. DPT effectively reduces the trainable parameters of PT. More details about baselines can be
found in Appendix B.
\subsection{Training Detail Settings}
\subsubsection{Implementation details}
The main experiments of EPT and baseline are performed using the T5-Base model \citep{shi2024dept}, which has a parameter size of 220M and the hidden size $d$ is 768. Consistent with the experimental setup of DEPT, we decompose the vanilla prompt (parameter size is 76,800) with the length of prompt tokens of 100. We train for 30,000 steps on small datasets with less than 100k training examples and 300,000 steps on large-size data with more than 100k examples. We conduct a grid search for learning rates and batch size is 16. the number of spaces is 4. Following DEPT, we utilize five source tasks - MNLI, QQP, SST-2, SQuAD, and ReCoRD - for the few-shot experiments. We derive our soft prompt from one of these selected source tasks to initialize our soft prompt and low-rank matrices. See Appendix C for the full experimental setup details. 

\subsubsection{Models}
Our models for evaluating EPT performance are T5-Small (60M), T5-Base (220M), T5-Large (770M), T5-3B, T5-11B and Llama2-7B \citep{touvron2023llama}. In this context, we employed quantization techniques when using T5-3B, T5-11B and Llama2-7B. PT performs poorly in smaller models, varying significantly based on hyperparameter selection \citep{vu2022spot}. Therefore, our primary experimental analysis focuses on the T5-base model.

\subsection{ Overall Performance Comparison (RQ1)}
Overall, Table \ref{tab_glue} show the results of EPT and other baselines on the GLUE and SuperGLUE benchmarks, respectively. Overall, EPT utilizes only a tiny number of trainable parameters yet consistently delivers outstanding performance across various downstream tasks. Additionally, it surpasses 11 other PEFT methods in average performance on two benchmarks, including PT variants based on multitasking and transfer learning. The visualized results of baselines are shown in Figure. \ref{Figure.1} and details in Appendix D. 

Among all baselines, although the full fine-tuning performs best in some datasets (MNLI, QQP, SST-2, and SuperGLUE\_Wic), the number of parameters required for training is 2,904 times that EPT, making full fine-tuning undoubtedly very computationally resource intensive. SPoT, DEPT, and EPT perform better while keeping the same training parameters as the original PT. This proves that randomly sampled tokens from the vocabulary for initialization and then directly injecting them into PLMs cannot make PT well adaptable to different downstream tasks. EPT and DEPT also utilize decomposing the soft prompt to reduce computing resources. 
Additionally, compared to the baseline MPT and ATTEMPT, which are the best-performing transfer learning methods, EPT performs better. EPT not only does not require additional pre-training source tasks but also trains fewer parameters.

Unlike SPoT and ATTEMPT, EPT has consistent performance in downstream tasks with different requirements, whereas they all utilize the attention mechanism. Additionally, SPoT and ATTEMPT only consider the relationship between source prompts of different tasks. EPT enhances the short prompt's semantic knowledge through the prompt fusion module and improves its adaptability to downstream tasks with different requirements by reweighting the short prompt in the multi-space projection module, which is why it performs better than EPT et al. Full fine-tuning performs best in some datasets, such as MNLI and QQP. We analyze that EPT is more efficient in datasets with fewer training samples. Overall, in the GLUE benchmark, our optimal baseline DEPT is only 0.3\% higher than MPT in single-task setting. DEPT is only 0.5\% better than MPT on multi-task setting on the SuperGLUE benchmark. On the contrary, on the GLUE benchmark, our proposed EPT outperforms DEPT by 1.5\% and vanilla PT by 2.9\%. On the SuperGLUE benchmark, EPT outperforms DEPT by a relative 1.7\% and vanilla PT by a relative 12.9\%. Therefore, while training time decreased by 14\%, the degree of performance improvement is already very noticeable.

\begin{table}[t]
\centering
\renewcommand{\arraystretch}{.7}
\setlength{\tabcolsep}{0.06cm}
    \begin{tabular}{ccc|c|>{\columncolor{gray!25}}c}
        \hline
        Prompt   & Prompt & Multi-Space & GLUE & Super-\\
        Decomposition & Fusion & Projection & (\%)  & GLUE (\%)\\
        \hline
        \XSolidBrush &\XSolidBrush &\XSolidBrush &84.3 & 67.3 \\
        \Checkmark  &\XSolidBrush &\XSolidBrush &85.8    & 76.3 \\\
        \Checkmark  &\XSolidBrush &\Checkmark &86.4     & 77.1 \\
        \Checkmark  &\Checkmark &\XSolidBrush &86.5    & 76.8 \\
        \Checkmark  &\Checkmark &\Checkmark &86.8     & 77.5 \\
        \hline
    \end{tabular}
    \caption{Performance comparison on the critical components of EPT on GLUE and SuperGLUE benchmarks.}
    \label{tab_ablation}
\end{table}

\begin{figure}[b]
\centering
  \includegraphics[scale=0.235, trim={0mm 0mm 0mm 0mm}]{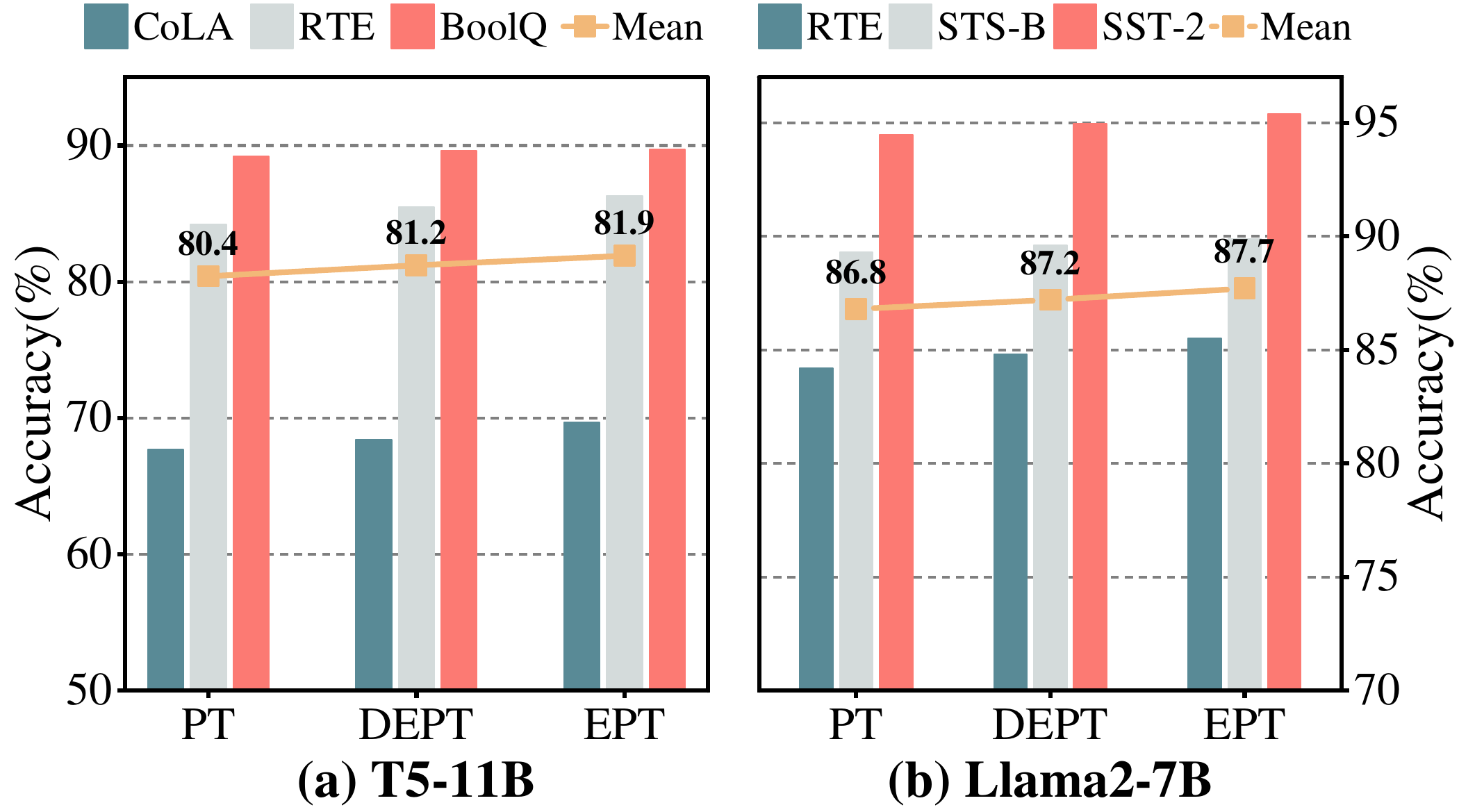}
    \caption{The performance changes of EPT(Ours), DEPT, and PT at different datasets on the T5-11B and Llama2-7B.}
    \label{Fig_11b_llama}
\end{figure}
\subsection{Ablation Experiment Analysis (RQ2)}
\subsubsection{Analysis the Critical Components of EPT}
To verify the contribution of each critical component (Prompt Decomposition, Prompt Fusion, and Multi-Space Projection) in EPT. 
We divided EPT into five different variants for ablation experiments, as shown in Table \ref{tab_ablation}. Overall, the result of EPT considering all critical components (i.e., the last line) is the most outstanding. The lack of any critical component in EPT significantly reduces performance, proving that each critical component positively impacts EPT. 
When not considering all critical components (i.e., the first line), EPT is a vanilla PT. When using the prompt fusion or multi-space projection module, EPT is superior to only performing the prompt decomposition. This again proves the effectiveness of the prompt fusion and multi-space projection module. 
\subsubsection{Power of Model Scale}
We conducted an empirical analysis of the impact of model size on performance using different datasets, as detailed in Table \ref{tab_T5-3B} (T5-3B) and Figure \ref{Fig_11b_llama} (T5-11B and Llama2-7B). We choose baselines initialized from a sampled vocabulary for comparison. As illustrated in Table \ref{tab_T5-3B} and Figure. \ref{Fig_11b_llama}, EPT outperforms other baselines across various datasets, with an average performance increase of 5.6\% on T5-3B compared to the original PT; this advantage persists even in larger models (T5-11B and Llama2-7B). Notably, all methods perform well in larger model scales, resulting in less pronounced performance differences, aligning with previous research findings \citep{lester2021power}. EPT is also capable of adapting to various downstream tasks in different model architectures.
Detailed performance comparisons of different baselines on T5-small, T5-Base, and T5-large are presented in Appendix E.

\begin{table}[t]
\renewcommand{\arraystretch}{0.7}
\setlength{\tabcolsep}{0.3cm}
  \centering
    \begin{tabular}{ll|ccc}
    \toprule
    \multicolumn{2}{l|}{\textbf{Model}} & \multicolumn{1}{c}{PT} & \multicolumn{1}{c}{DEPT} & \multicolumn{1}{c}{EPT} \\
    \midrule
    BoolQ & (9.4K) & 87.0  & 87.8  & 87.9  \\
    CoLA  & (8.5K) & 66.1  & 67.8  & 68.2  \\
    WiC   & (6.0K) & 70.5  & 71.2  & 73.7  \\
    MultiRC & (5.1K) & 78.0  & 80.5  & 80.8  \\
    MRPC  & (3.7K) & 90.7  & 91.7  & 92.2  \\
    RTE   & (2.5K) & 82.7  & 84.2  & 85.6  \\
    WSC   & (554) & 67.3  & 67.3  & 69.2  \\
    CB    & (250) & 75.0  & 94.6  & 94.6  \\
    \midrule
    \rowcolor{gray!25}
    Mean  & (\%)  & 77.2  & 80.6  & 81.5  \\
    \bottomrule
    \end{tabular}%
    \caption{Performance comparison of PT, DEPT and EPT on different datasets for T5-3B.}
  \label{tab_T5-3B}%
\end{table}%


\subsection{Indepth Analysis (RQ3)}
\subsubsection{Few-shot adaptation}
Following previous work \citep{asai2022attempt,wang2022multitask,shi2024dept}, we pre-trained the soft prompt and the low-rank matrices on source tasks. We evaluate the performance of EPT, vanilla PT, and MPT in $k$-shot (k = 4, 16, 32) on the GLUE benchmark. As shown in Figure. \ref{Figure.4}(a), the performance improvement of EPT is mainly due to using the PETL framework for pre-training source prompts. EPT outperforms other variants of PT under few-shot learning tasks, which proves its effectiveness.
\subsubsection{The Length of Soft Prompt}
\begin{figure}[t]
\centering
  \includegraphics[scale=0.212, trim={0mm 0mm 0mm 0mm}]{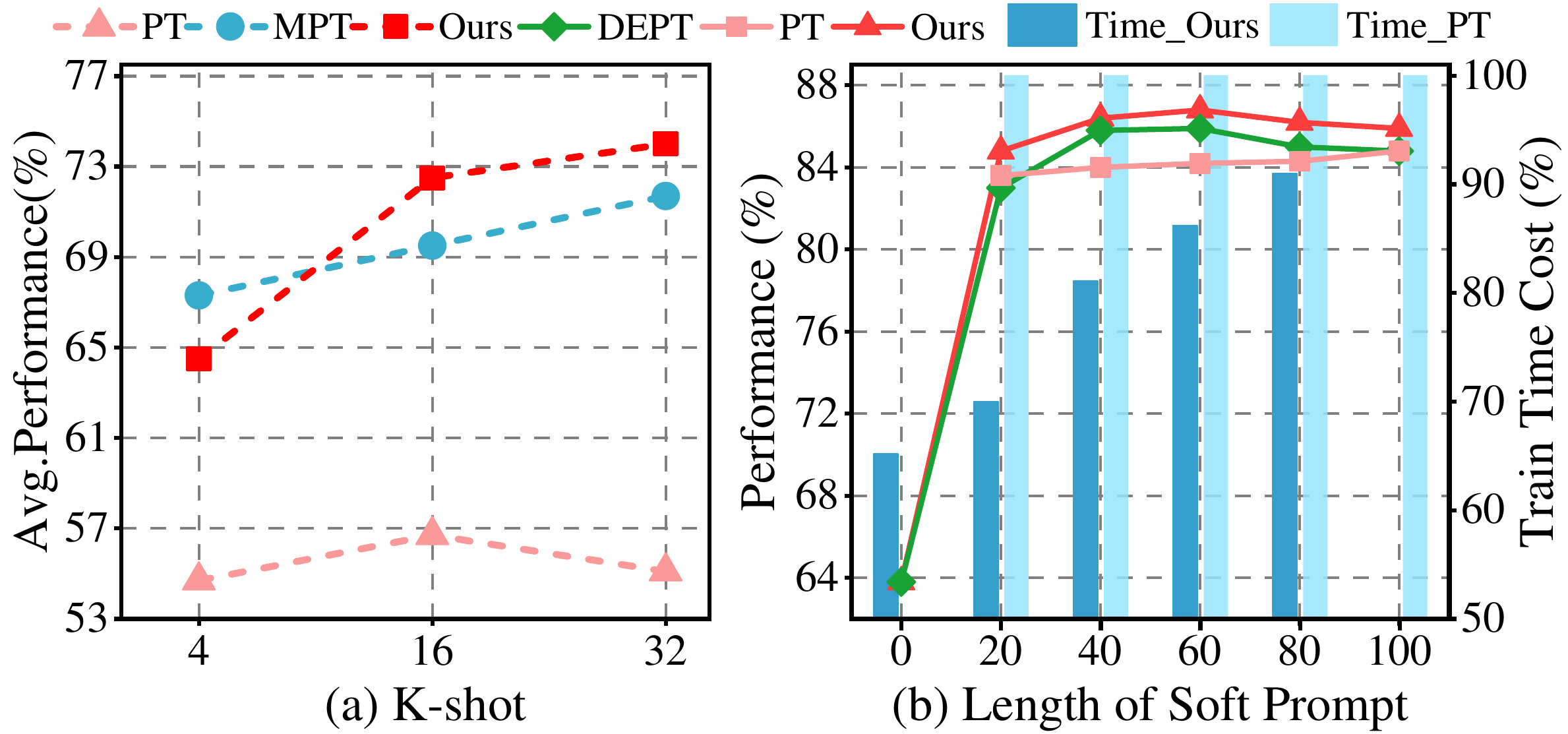}
    \caption{On the GLUE benchmark, (a) The performance changes of EPT(Ours), MPT, and PT at different K-shot. (b) Comparison of training time consumption and the performance changes (EPT, DEPT, and PT) according to different lengths of the short prompt in EPT and DEPT.}
    \label{Figure.4}
\end{figure}

For the EPT method, we maintained the same number of trainable parameters (76,800) as the conventional PT with a length of 100, and compared the training time costs between EPT and PT. Figure. \ref{Figure.4}(b) shows that EPT takes more training time as the length of the short prompt increases. When the length of the short prompt is 60, EPT has the best performance on the GLUE  benchmark, and the training time of EPT is 14\% lower than that of the original PT. On the GLUE benchmark, EPT significantly outperforms DEPT and PT at different prompt's lengths (except for length 0). When the length is 0, the source prompt is only decomposed into two low-rank matrices, rendering the prompt fusion and multi-space projection modules in EPT non-functional. Consequently, EPT and DEPT exhibit identical performance. Additionally, the parameters of vanilla PT are frozen and not updated, resulting in no performance outcomes. When the soft prompt length is 100, DEPT is conventional PT, and EPT outperforms DEPT as the short prompts are mapped to different subspaces to reweight the prompt tokens, positively influencing EPT. This demonstrates that conventional PT struggles to adapt to downstream tasks with varying requirements through fine-tuning in the same single embedding space. Due to the GLUE and SuperGLUE benchmarks, which include many datasets, using average performance to compare improvement rates may create an illusion of non-significant parameter influence. Hence, the detailed changes in EPT' performance in terms of soft prompt length in Appendix F.

\subsubsection{The Impact of the Number of Spaces }
\begin{figure}[t]
\centering
  \includegraphics[scale=0.185, trim={0mm 0mm 0mm 0mm}]{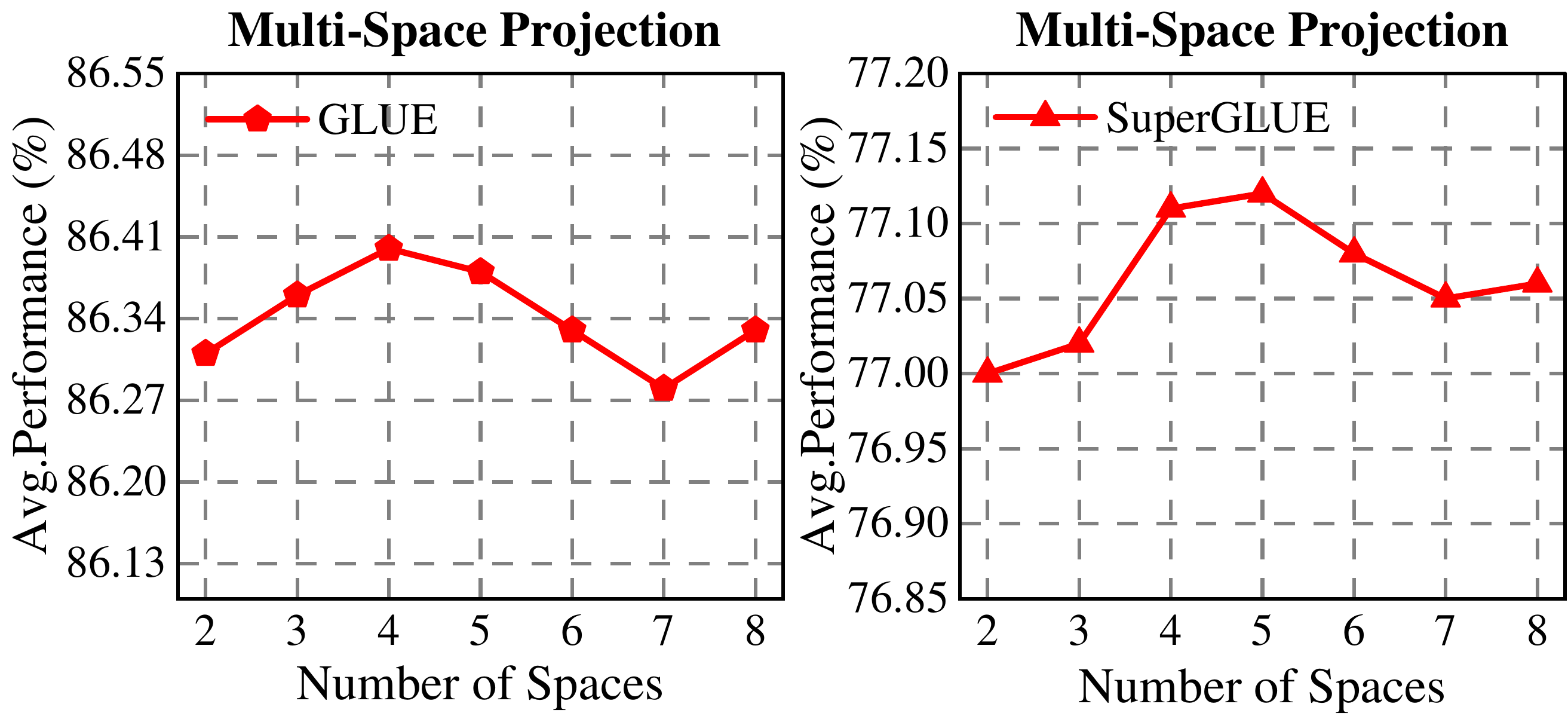}
    \caption{Performance of the number of spaces in the Multi-Space Projection module on the GLUE and SuperGLUE benchmarks.}
    \label{Figure.5}
\end{figure}

To eliminate the noise generated by the prompt fusion module, when analyzing the impact of changes in the number of spaces on performance, we only leverage a multi-space projection module that learns the reweighted short prompt. As shown in Figure. \ref{Figure.5}, we dynamically alter the number of spaces $N$ from 2 to 8 with a step size of 1 during training. Overall, there are many datasets in both the GLUE and SuperGLUE benchmarks, so the fluctuations in EPT on the two benchmarks are small, and the number of spaces we comprehensively selected is 4. We also visualized the weight allocation of the gating network to different prompt tokens in Appendix G.

\section{Related Works}
\subsection{Parameter-efficient Fine-tuning}
Parameter-efficient fine-tuning approaches can adapt well to various downstream tasks by updating a limited number of training parameters compared to full fine-tuning. 
AdapterDrop \citep{ruckle2021adapterdrop} dynamically dropping the Adapter reduces the number of model parameters as much as possible and improves the efficiency of model training/inference. Diff pruning \citep{guo2021parameter} learns a task-specific ``diff'' vector that extends the original pre-trained parameters. LoRA \citep{hu2021lora} only updates the parameters of low-rank matrix pairs. BitFit \citep{zaken2022bitfit} only updates the mask layer parameters of PLMs. HyperDecoder \citep{ivison2022hyperdecoders} efficient adaptation of parameters for decoder generation using a hyper-network conditioned on encoder output in multi-task. 
LST \citep{sung2022lst} aims to reduce the training memory by a ladder-side network for transformers. Prompt tuning (PT) is a promising parameter-efficient fine-tuning (PEFT) approach, as its parameters do not exhibit dramatic growth even when the model size expands significantly.
\ \\ \noindent
\subsection{PT-based methods}
The expansion in PLMs size does not lead to a surge in the training parameters of PT. The recent research aims to improve the performance of PT through various approaches. SPoT \citep{vu2022spot} learns one or more source prompts, constructing the interaction with the target task to initialize the target prompt. 
ATTEMPT \citep{asai2022attempt} considers the impact of knowledge in the source prompts on the input sequence to generate different attention weights, achieving weighting target prompts. MPT \citep{wang2022multitask} decomposes each source prompt into a one-rank matrix, performs Hadamard product with shared prompts to construct student prompts, and then improves the performance of PT through knowledge distillation. 
DPT \citep{xiao2023decomposed} initializes a soft prompt to reduce the number of trainable parameters by utilizing two low rank vectors instead of soft prompt. 
These variants, which are built upon soft prompts, have exhibited remarkable performance. However, these PT-based methods still struggle to balance efficiency and accuracy. Moreover, they typically work in a single space, thus resulting in performance inconsistencies across different downstream tasks.

\section{Conclusions and Future Work}
In this work, we propose an efficient soft prompt tuning (EPT) method by prompt fusion and multi-space projection. Specifically, the prompt fusion module can help enhance the semantic of the soft prompt, leading to a balance between accuracy and efficiency. The multi-space module projects a single soft prompt into multiple subspaces with reweighted prompt tokens, improving the performance consistency. Experimental results across two model architectures (T5 and Llama2) demonstrate that EPT reduces training time, achieves optimal and consistent performance using the shorter soft prompt, and validates the effectiveness of critical components in EPT.

For future work, we will address the computational overhead introduced by using two learning rates in EPT for parameter search. Furthermore, we intend to explore the integration of EPT with soft prompt methods based on multi-task transfer learning, aiming to reduce training parameters further while maintaining optimal performance. 

\section{Acknowledge}
This work is partially supported by the National Natural Science Foundation of China under Grant No. 62032013, the Science and technology projects in Liaoning Province (No. 2023JH3/10200005), and the Fundamental Research Funds for the Central Universities under Grant No. N2317002.

\bibliography{aaai24}

\section{Appendices}

\subsection{Appendix A: Datasets Details}
\label{Datasets Details}
Our experimental datasets include eight datasets from the GLUE benchmark
\citep{wang2018glue}, six datasets from the SuperGLUE benchmark 
, and one dataset \footnote{\footnotesize \url{https://huggingface.co/lucadiliello}} from the MRQA 2019 Shared Task \cite{wang2019superglue}. Table~\ref{tab_benchmark} shows the complete statistics of all experimental datasets, including the size of training, validation, and test sets, the task type, the domain to which each dataset belongs, and the evaluation metrics. 

\begin{table*}[t]
  \centering
  \renewcommand{\arraystretch}{1.1}
  \setlength{\tabcolsep}{0.12cm}
    \begin{tabular}{lrrrlll}
    \toprule
    \multicolumn{7}{c}{\textbf{GLUE Benchmark}} \\
    \midrule
    \textbf{Dataset} & \multicolumn{1}{l}{\textbf{\#Train}} & \multicolumn{1}{l}{\textbf{\#Valid}} & \multicolumn{1}{l}{\textbf{\#Test}} & \textbf{Type} & \textbf{Domain} & \textbf{\#Metric} \\
    \midrule
    MNLI  & 392,702 & 9,832 & 9,815 & NLI   & various & accuracy  \\
    QQP   & 362,846 & 1,000 & 40,431 & Paraphrase & social QA questions (Quora) &
    accuracy
    \\
    QNLI  & 103,743 & 1,000 & 5,463 & NLI   & Wikipedia & accuracy  \\
    MRPC  & 3,668 & 204   & 204   & Paraphrase & news  & 
    accuracy
    \\
    STS-B & 5,749 & 750   & 750   &  Sent. Similarity & various &
    Pearson corr. 
    \\
    SST-2 & 66,349 & 1,000 & 872   & Sentiment & Movie Reviews & accuracy  \\
    CoLA  & 8,551 & 521   & 522   & Acceptability & various & Matthews corr. \\
    RTE   & 2,490 & 138   & 139   & NLI   & News, Wikipedia & accuracy  \\
    \midrule
    \multicolumn{7}{c}{\textbf{SuperGLUE Benchmark}} \\
    \midrule
    \textbf{Dataset} & \multicolumn{1}{l}{\textbf{\#Train}} & \multicolumn{1}{l}{\textbf{\#Valid}} & \multicolumn{1}{l}{\textbf{\#Test}} & \textbf{Type} & \textbf{Domain} & \textbf{\#Metric} \\
    \midrule
    MulticRC & 27,243 & 2,424 & 2,424 & Question Answering & various & F1
    \\
    Wic   & 5,428 & 319   & 319   & Word Sense Disambiguation & lexical databases & accuracy  \\
    WSC   & 554   & 52    & 52    & Common Sense Reasoning & fiction books & accuracy  \\
    BoolQ & 9,427 & 1,635 & 1,635 & Question Answering & Wikipedia & accuracy  \\
    CB    & 250   & 28    & 28    &  NLI  & various & accuracy  \\
    ReCoRD & 137,484 & 1,370 & 15,176 & Common Sense Reasoning & news (CNN, Daily Mail) & 
    F1
    \\
    \midrule
    \multicolumn{7}{c}{\textbf{MRQA 2019 Shared Task}} \\
    \midrule
    SQuAD & 87,599 & 10,570 & -  & Question Answering & Wikipedia & 
    F1
    \\
    \bottomrule
    \end{tabular}%
    \caption{The details of the 15 datasets used in our experiment. NLI stands for natural language inference.}
  \label{tab_benchmark}%
\end{table*}%

\subsection{Appendix B: Datasets Details}
Descriptions of all baselines are as follows:
\begin{itemize}
\item \textbf{Fine-tuning:} Updating all model parameters in the PLM on each downstream task.
\item \textbf{Adapter} ~\citep{houlsby2019parameter}: A parameter-efficient method adds an Adapter module to some layers of the pre-training model, and the Adapter module learns the knowledge of specific downstream tasks.
\item \textbf{AdapterDrop} ~\citep{ruckle2021adapterdrop}: Dynamically dropping the Adapter reduces the number of model parameters as much as possible and improves the efficiency of model training/inference.
\item \textbf{AdaMix} \citep{wang2022adamix}: trained with stochastic routing and adaptation module merging to maintain computational cost, serves as a general PEFT method by tuning a mixture of adaptation modules introduced in each Transformer layer while keeping most of the PLM weights frozen.
\item \textbf{BitFit} ~\citep{zaken2022bitfit}: An efficient fine-tuning method for updating the mask layer parameters of PLM to adapter downstream tasks.
\item \textbf{LoRA} \citep{hu2021lora}: A parameter-efficient method only updates the parameters of low-rank matrices.
\item \textbf{PT} \citep{lester2021power}: Updating the parameters within the soft prompts added to the model's input embedding layer to accommodate various downstream tasks.
\item \textbf{SPoT} \citep{vu2022spot}: A novel prompt-based transfer learning approach acquires one or more source prompts and establishes interactions with the target task to initialize the target prompt.
\item \textbf{ATTEMPT} \citep{asai2022attempt}: Considering the prompts of different tasks and the weight relationship between the target prompt and input.
\item \textbf{MPT} \citep{wang2022multitask}: Decomposing prompt into shared prompts and low-rank matrices and using prompt distillation to make the model more suitable for downstream tasks.
\item \textbf{DEPT} \citep{shi2024dept}: Decomposing  the soft prompt into smaller prompts and low-rank matrix pairs to reduce training time.
\item \textbf{DPT} \citep{xiao2023decomposed}: Multiplying two randomly initialized low-rank matrices serves as a new approach for prompt length initialization.
\end{itemize}

\subsection{Appendix C: Training Detail Settings}
Our implementation is based on PyTorch 1.13.1 \footnote{\footnotesize \url{https://pytorch.org/}}, Huggingface Transformers 4.41.0 \footnote{\footnotesize \url{https://github.com/huggingface/transformers}}, and Huggingface PEFT 0.21.4 \footnote{\footnotesize \url{https://github.com/huggingface/peft}}. All of our experiments were conducted with 8 GPUs, with 48 GB memory each. Table~\ref{tab_params} gives an exhaustive enumeration of the hyper-parameters used in the experiments. Specifically, we use different learning rates for decomposed soft prompts and low-rank matrices. For soft prompts, we search for learning rate within the set \{3e-1, 4e-1, 5e-1\}; for the low-rank matrices, we search for learning rate within the set \{1e-04, 5e-4, 5e-03\}. Finally, the maximum sequence length of the model is usually set to 256 (the length on SuperGLUE-MultiRC is set to 348), and we evaluate performance every 1000 steps. 

\begin{table}[h]
  \centering
  \renewcommand{\arraystretch}{1.1}
    \begin{tabular}{l|l}
    \toprule
    \textbf{Hyperparameter} & \multicolumn{1}{l}{\textbf{Assignment}} \\
    \midrule
    number of steps & 30,000 steps \\
    batch size & 16 \\
    maximum learning rate ($\alpha_{1}$) & 3e-1, 4e-1, 5e-1 \\
    maximum learning rate ($\alpha_{2}$) & 1e-04, 5e-04, 5e-03 \\
    number of spaces & 2, 3, 4, 5, 6, 7, 8 \\
    length of the soft prompt (s) & 20, 40, 60, 80 \\
    maximum sequence length & 256 \\
    \bottomrule
    \end{tabular}%
    \caption{Hyper-parameters for our EPT.}
  \label{tab_params}%
\end{table}%

\subsection{Appendix D: Accuracy vs. efficiency}
Figure. 1 visualizes the relationship between all baseline performance and the amount of training parameters on the GLUE and SuperGLUE benchmarks. EPT utilize the efficient joint representation of prompts to displace the original prompt, and the trainable parameters input to PLMs remain consistent with ordinary prompts. We experimentally demonstrate that EPT training with fewer parameters, outstanding performance, and strong adaptability to different downstream tasks.

\begin{figure}[t]
\centering
  \includegraphics[scale=0.37, trim={0mm 0mm 0mm 0mm}]{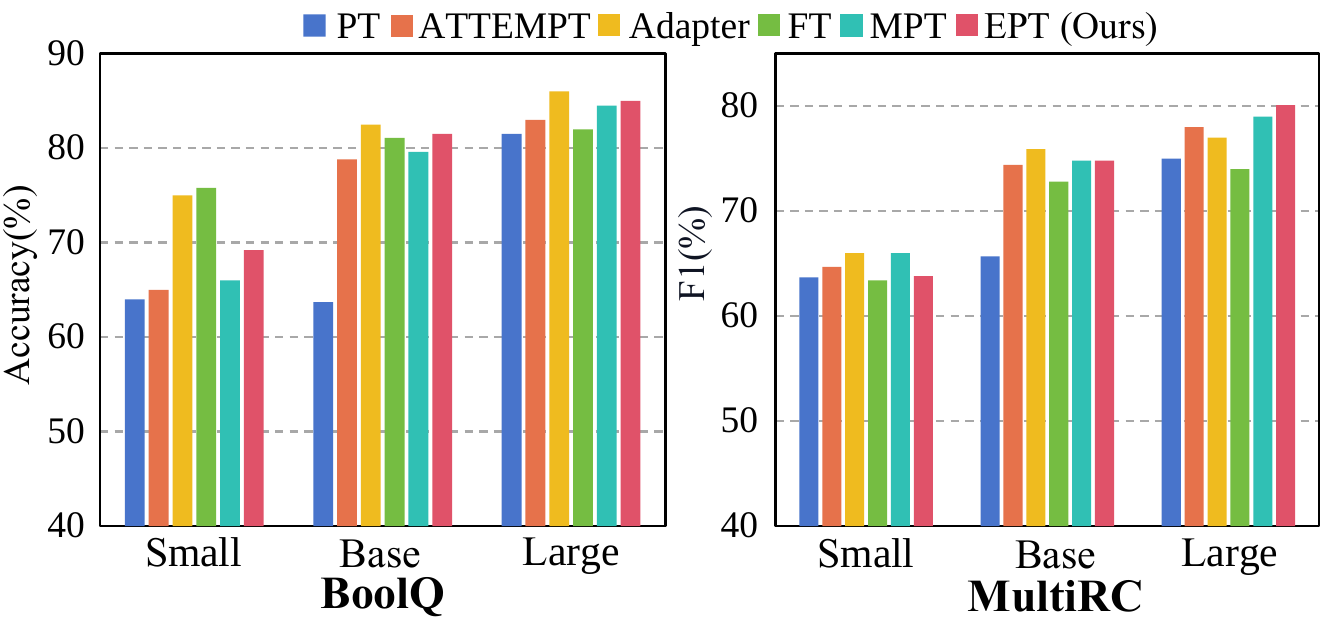}
    \caption{Performance of different baselines varies with model scale (from T5-Small, T5-Base, T5-Large).}
    \label{Figure.3}
\end{figure}
\subsection{Appendix E: Power of Model Scale} 
We empirically analyze the impact of the model scale (T5-Small, T5-Base, T5-Large) on the performance of different baselines (PT, ATTEMPT, Adapter, Fine-tuning, MPT, and EPT) on BoolQ and MultiRC in the SuperGLUE benchmark. As shown in Figure.\ref{Figure.3},  the performance get improvements with the increase of the model scale. This is aligned with the findings of ~\citep{lester2021power}. When the model specification is T5-Large, the performance of EPT on MultiRC also is optimal. Furthermore, the performance of EPT outperforms the other baselines on the SuperGLUE benchmark with T5-Large, and the parameters required to train EPT are far less than full fine-tuning and Adapter.

\subsection{Appendix F: Details of EPT Performance}
\begin{figure}[t]
\centering
  \subfloat{
   \includegraphics[scale=0.21, trim={0mm 0mm 0mm 0mm}]{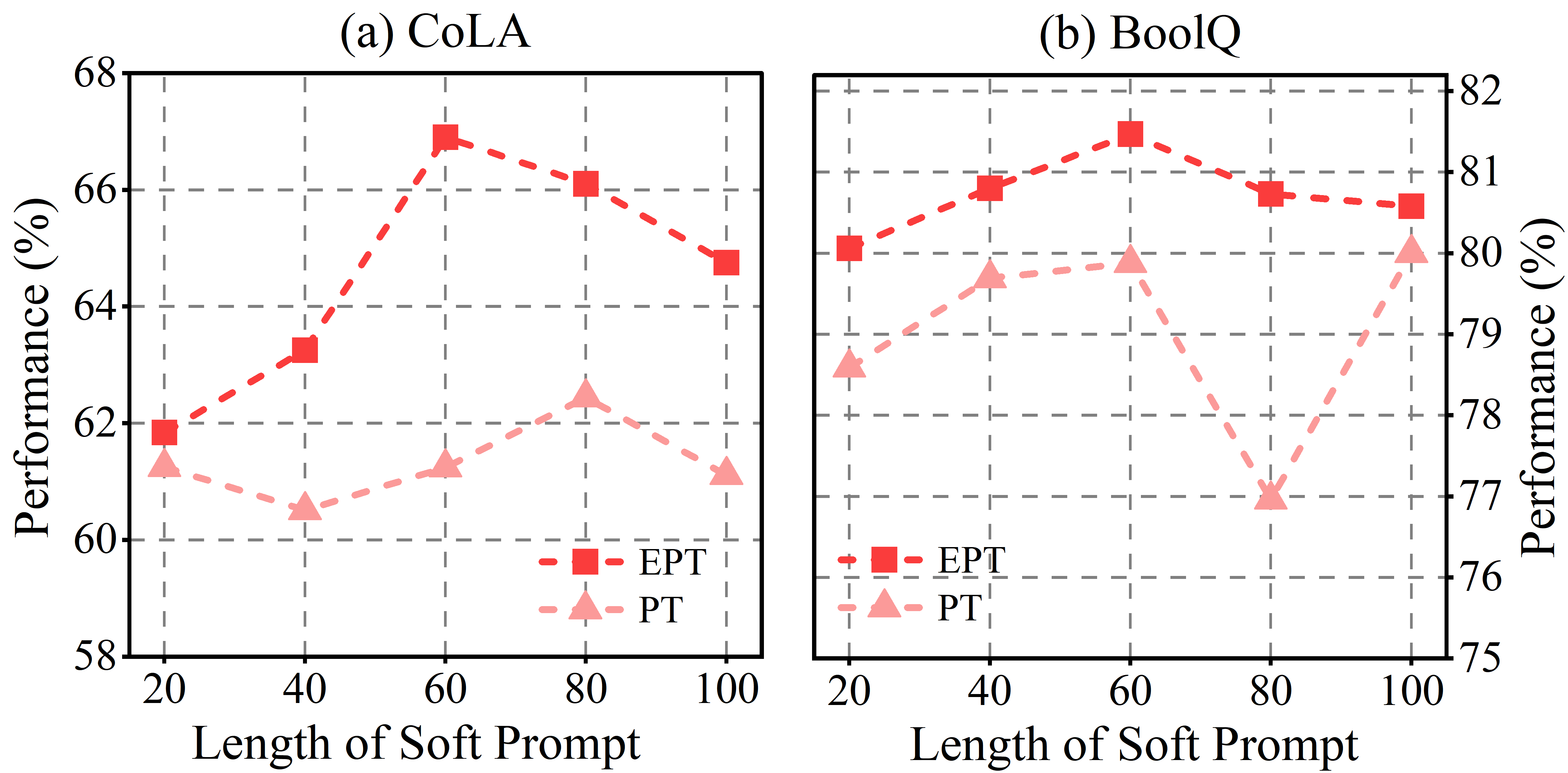}}
  \vspace{0.5pt}
    \caption{(a) and (b) are the performance comparisons of PT and EPT as the length changes on the CoLA and BoolQ, respectively.}
    \label{Figure_cola}
\end{figure}

Given that the GLUE and SuperGLUE benchmarks encompass a multitude of datasets, reliance on aggregate performance metrics for assessing improvement rates might obscure the actual impact of parameter modifications. As shown in Figure. \ref{Figure_cola}(a) and Figure. \ref{Figure_cola}(b), To display the performance of EPT more intuitively, we randomly compared the performance of vanilla PT and EPT on the CoLA and BoolQ datasets. EPT performs better than PT under different prompt lengths.

\begin{figure}[t]
\centering
    \subfloat{
		\includegraphics[scale=0.029]{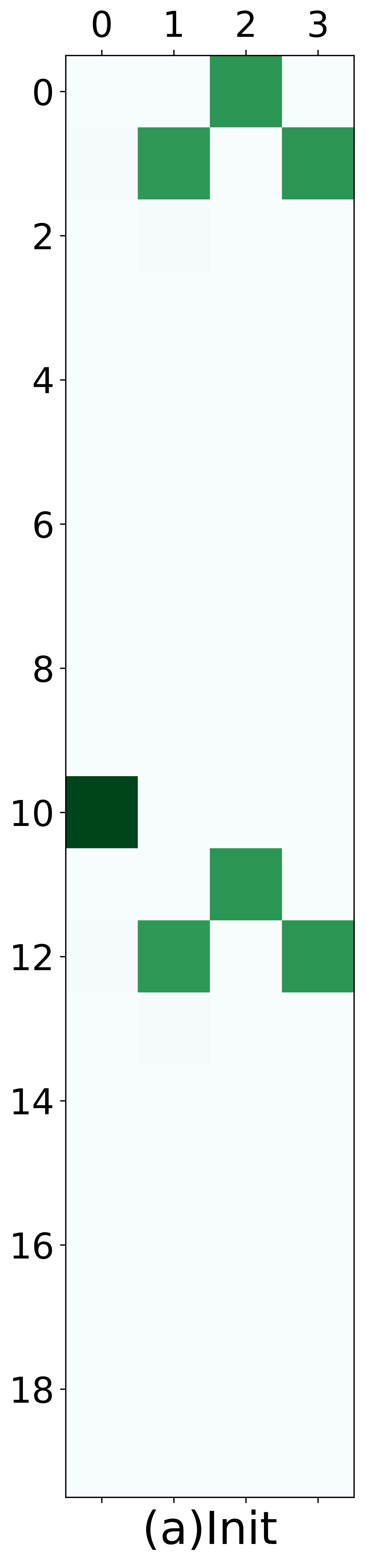}
   }
     \subfloat{
		\includegraphics[scale=0.029]{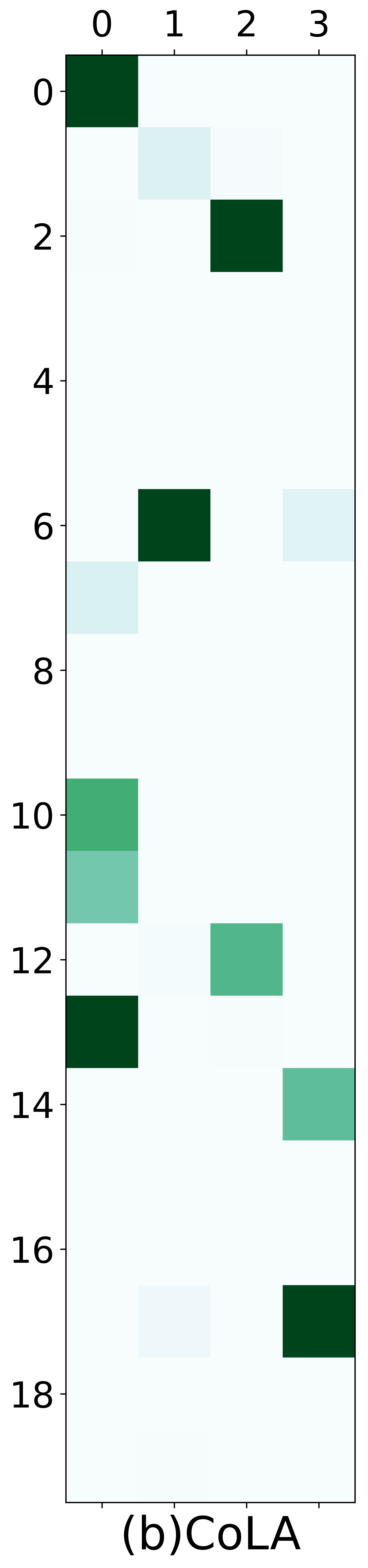}
  }
    \subfloat{
		\includegraphics[scale=0.029]{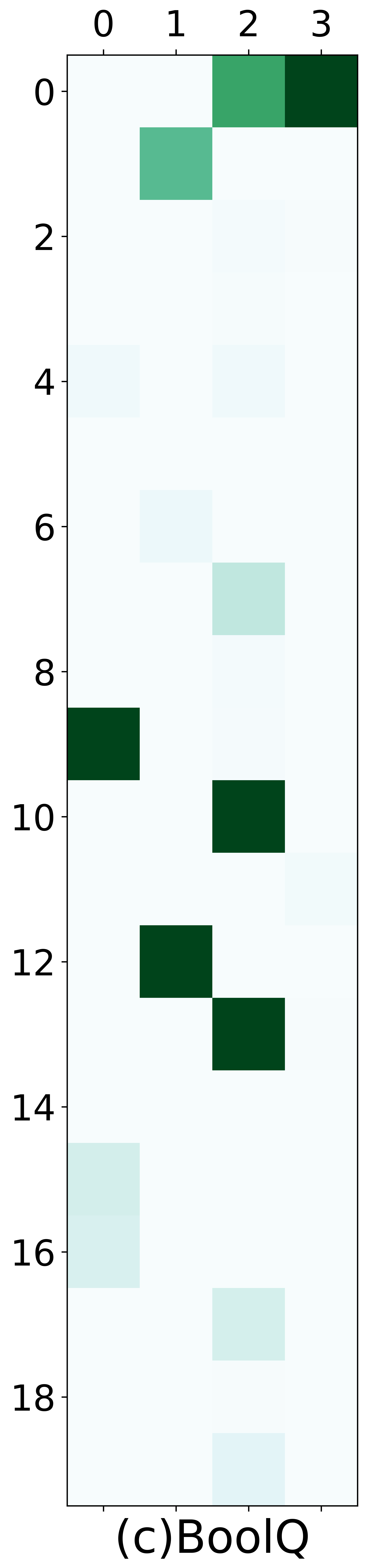}
  }
  \subfloat{
		\includegraphics[scale=0.029]{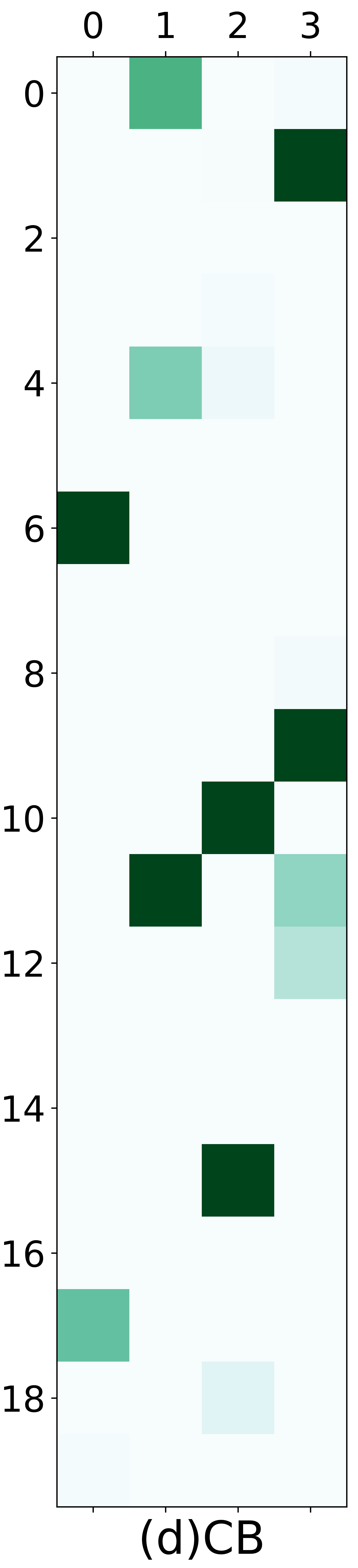}
  }
  \subfloat{
		\includegraphics[scale=0.029]{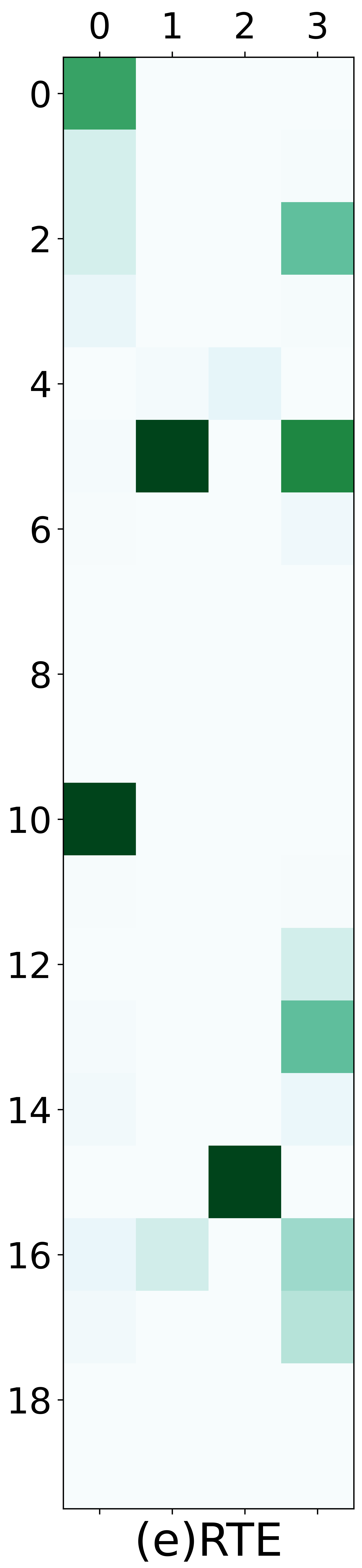}
  }
\caption{The short prompt (length is 20) maps into four different subspaces and the attention level of each short prompt token is compared with initialization (derived from an existing vocabulary) on four datasets (i.e., CoLA, BoolQ, CB, and RTE).}
    \label{fig_weight}
\end{figure}
\subsection{Appendix G: Interpretability of Weights in Spaces}
As shown in Figure. \ref{fig_weight}, after mapping the short prompt into distinct spaces, the distribution of weights across various prompt tokens by the gating network was elucidated through visualization. We are comparing the weight distribution of the original prompt tokens initialized in the existing vocabulary on four datasets (i.e., CoLA, BoolQ, CB, and RTE). It is evident that different downstream tasks exhibit distinct attention levels towards tokens in prompts, which proves that reweighting short prompt tokens in the multi-space projection module is indispensable for improving the stability of PT.

\end{document}